\documentclass{article}
\usepackage{iclr2027_conference,times}
\usepackage{amsmath,amsfonts,bm}

\def\eqref#1{equation~\ref{#1}}
\def\1{\bm{1}}

\DeclareMathAlphabet{\mathsfit}{\encodingdefault}{\sfdefault}{m}{sl}
\SetMathAlphabet{\mathsfit}{bold}{\encodingdefault}{\sfdefault}{bx}{n}

\usepackage{graphicx}
\graphicspath{{contens/figures/}}
\usepackage{booktabs}
\usepackage{array}
\usepackage{multirow}
\usepackage{wrapfig}
\usepackage{amsmath,amssymb}
\usepackage{xcolor}
\usepackage[hidelinks]{hyperref}
\usepackage{url}
\usepackage{xspace}

\newcommand{\method}{Agentic-OPD\xspace}

\newcommand{\student}{\pi_{S,t}}

\newif\ifarxivversion
\arxivversiontrue

\title{Agentic-OPD: Recovering Forgotten Agent Capabilities via On-Policy Distillation for Agentic Tasks}

\ifarxivversion
  \iclrfinalcopy
\fi

\newcommand{\cameraauthorblock}{%
\parbox{0.94\textwidth}{\centering
Tingcheng Bian$^{1,*}$ \quad
Kangye Gong$^{1,*}$ \quad
Jinchang Luo$^{1}$ \par
MingQuan Cheng$^{1}$ \quad
Feiran Liu$^{2}$ \quad
Ni Li$^{1}$ \par
Xiaoling Xia$^{1}$ \quad
Haiwei Wang$^{1}$ \quad
Hui Li$^{1,\dagger}$ \par
{\normalfont\footnotesize
$^{1}$Baidu, Inc. \quad $^{2}$Beijing Jiaotong University \par
$^{*}$Equal contribution. \quad
$^{\dagger}$Corresponding author: \texttt{lihui24@baidu.com}. \par
\texttt{biantingcheng@baidu.com},
\texttt{gongkangye@baidu.com},
\texttt{luojinchang@baidu.com} \par
\texttt{26110178@bjtu.edu.cn},
\texttt{lini03@baidu.com},
\texttt{xiaxiaoling@baidu.com}}
}}
\author{\cameraauthorblock}

\begin{document}
\maketitle

\ifarxivversion
  % 覆盖 style 在 final 模式下写入的 "Published as a conference paper at ICLR 2027"
  \lhead{Preprint}
  \rhead{}
  \hypersetup{
    pdftitle={Agentic-OPD: Recovering Forgotten Agent Capabilities via On-Policy Distillation for Agentic Tasks},
    pdfauthor={Tingcheng Bian, Kangye Gong, Jinchang Luo, MingQuan Cheng, Feiran Liu, Ni Li, Xiaoling Xia, Haiwei Wang, Hui Li},
    pdfsubject={Preprint},
    pdfkeywords={on-policy distillation, LLM agents, reinforcement learning, capability recovery}
  }
\else
  % 双盲版：PDF 元数据也不能泄露作者身份
  \hypersetup{
    pdftitle={Agentic-OPD: Recovering Forgotten Agent Capabilities via On-Policy Distillation for Agentic Tasks},
    pdfauthor={Anonymous},
    pdfsubject={Under review at ICLR 2027}
  }
\fi

\begin{abstract}
Reinforcement learning is now the mainstream paradigm
for training LLMs on agentic tasks,
yet standard practice keeps only the best-validation checkpoint
and discards the rest.
We show that this is lossy:
in GRPO training of LLM agents,
the best-validation checkpoint fails on problems
that earlier checkpoints from the same run had already solved,
and the union of problems solved across checkpoints
far exceeds any single one.
The discarded checkpoints thus form a teacher pool
obtained at no additional training cost,
which on-policy distillation (OPD) can mine
to recover the lost coverage.
Mining this pool raises two challenges:
i) which checkpoint to distill from,
as no checkpoint dominates the others
and complementarity to the student is problem-specific;
ii) where to trust it,
as even a well-chosen teacher is only locally superior,
so indiscriminate distillation overwrites behavior
the student has already mastered.
We propose \method, which treats the best-validation checkpoint
as the student and all saved checkpoints as frozen teacher candidates,
using the student's evolving capability gaps to guide both
teacher selection and supervision filtering:
the HCN policy (Highest Coverage Next)
dynamically selects teachers by weighted coverage of the student's unsolved problems,
while the LAD module (Locality-Aware Distillation)
confines transfer to failed student rollouts on problems where the teacher performs better.
Across five agentic task domains and 24 benchmarks and task subsets,
\method improves over its initialization checkpoint
by up to 7.4 points.
\method recasts a finished RL run as a reusable pool of policies
and establishes post-RL capability recovery as a stage
orthogonal to improvements in the RL algorithm itself.
\end{abstract}

\section{Introduction}

Reinforcement learning with verifiable outcome rewards has emerged as a prominent
approach for turning language models into agents that can reason, call tools, and
act over multiple turns \citep{xu2025llm,li2025review,shao2024deepseekmath}.
Such runs produce dozens of checkpoints. Reporting only the best validation
checkpoint leaves the capabilities of other checkpoints unmeasured. In agentic
RL, an update can introduce a new behavior while erasing an earlier one. Across five
agentic domains, we find that checkpoints from a single run collectively solve
far more problems than any one checkpoint, revealing complementary capabilities
along the training trajectory (Section~\ref{sec:preliminary}).

Parameter-space merging produces one
policy but can erase checkpoint-specific strengths, while separate inference
increases deployment cost
\citep{li2026temporal}. Multi-teacher on-policy distillation (OPD) offers an
alternative by consolidating these complementary capabilities into a single
student policy
\citep{agarwal2024policy,gu2024minillm}. Prior work studies OPD with multiple teachers
\citep{ma2026mopd,wang2026mad,li2604rethinking}. Here, the teachers are temporal
checkpoints from a single run, each of which may outperform the student on
selected problems despite lower overall performance. This raises two
challenges: \textbf{teacher selection}, choosing checkpoints that complement the
student, and \textbf{weak-to-strong distillation}, transferring locally useful
behaviors from globally weaker checkpoints to the student.

We propose \method, which organizes distillation around the student's
evolving capability gaps: whom to learn from and where to accept supervision.
With all teacher checkpoints frozen, \textbf{HCN} (Highest Coverage Next)
dynamically activates the candidate covering most of the student's current
failures, while \textbf{LAD} (Locality-Aware Distillation) confines supervision
to student-generated tokens in failed rollouts on problems where that
teacher performs better.
Across five agentic domains and 24 benchmarks and task subsets, \method delivers consistent gains
over the converged GRPO initialization, with gains of 2.6 to 7.4 points on
DeepSearch across three backbones.

Our contributions are:
\begin{itemize}
    \item We quantify capability forgetting across five agentic domains,
    showing that checkpoints from a single RL run collectively solve far more
    problems than its best checkpoint.

    \item We formulate temporal checkpoint consolidation as weak-to-strong OPD:
    HCN dynamically routes teachers according to the student's capability gaps,
    while LAD transfers supervision only where a globally weaker teacher is
    locally stronger.

    \item We show that \method further improves performance after GRPO training
    has plateaued. We validate these gains across five agentic domains and on
    three backbones for DeepSearch.
\end{itemize}

\begin{figure}[t]
    \centering
    \begin{minipage}[t]{0.49\linewidth}
        \centering
        \includegraphics[width=\linewidth]{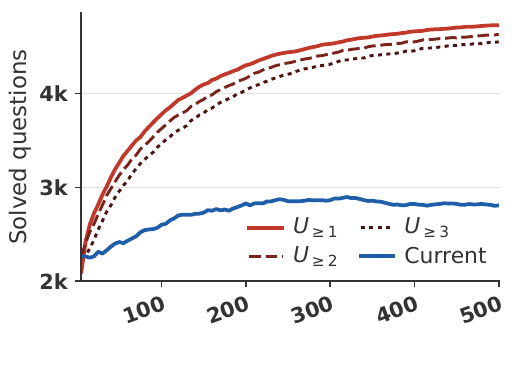}\\[2pt]
        {\small (a) DeepSearch (HotpotQA)}
    \end{minipage}\hfill
    \begin{minipage}[t]{0.49\linewidth}
        \centering
        \includegraphics[width=\linewidth]{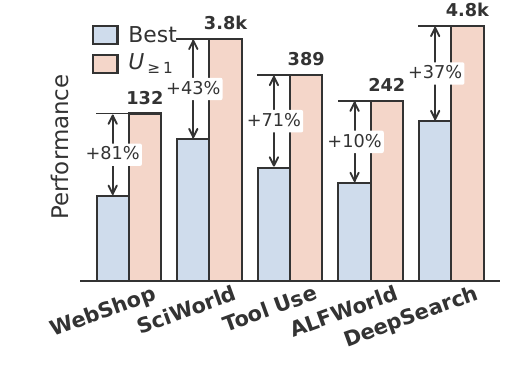}\\[2pt]
        {\small (b) Best checkpoint vs.\ $U_{\geq 1}$}
    \end{minipage}
    \caption{The temporal capability gap. \textbf{(a)} DeepSearch on HotpotQA:
    the current checkpoint against $U_{\geq k}(i)$, the tasks solved by at least
    $k$ of the checkpoints through index $i$, for $k=1,2,3$. \textbf{(b)} Best
    single checkpoint against the historical union $U_{\geq 1}$ in five domains,
    with the relative gap annotated.}
    \label{fig:preliminary-gap}
\end{figure}
\section{Preliminary Study}
\label{sec:preliminary}

An agent is a policy $\pi$ that interacts with an external environment over
multiple turns, interleaving reasoning, tool calls, and answer generation. We
score only the outcome: for a task $x$ with reference answer $y^*(x)$, let
$\hat y_{\pi}(x)$ be its final answer and
\begin{equation}
    C_{\pi}(x) = \mathbf{1}\!\left[\hat y_{\pi}(x)=y^*(x)\right],
    \label{eq:preliminary-correctness}
\end{equation}
substituting the environment's own success predicate where the outcome is not an
answer string.

\citet{li2026temporal} report that CoT-style mathematical reasoning forgets
intermediate abilities under both SFT and RL, with Temporal Forgetting Scores of
6.4\% to 56.1\%. We ask whether similar forgetting occurs in agentic reasoning,
which involves multi-turn interaction with external environments.

We measure forgetting on the five domains of Section~\ref{sec:exp-settings},
training each from Qwen2.5-3B-Instruct with GRPO under result-based rewards and
retaining every checkpoint of that run. Within each domain, all checkpoints
share the same base model, data mixture, and tool environment. DeepSearch is
the detailed case: we train on HotpotQA for $500$ steps and save and evaluate a
checkpoint every $5$ steps, yielding $n=100$ checkpoints. This run trains on
HotpotQA only and is separate from the NQ and HotpotQA runs used for
consolidation in Section~\ref{sec:experiments}. The held-out validation
set $\mathcal{D}_{\rm val}$ consists of a random $10\%$ split of the training
queries. We use greedy decoding for all checkpoint evaluations. The other four
domains follow the protocols of Section~\ref{sec:exp-settings}.
Let $\mathcal{D}_{\rm val}^{(d)}$ denote the validation set for domain $d$.
Writing $T_i$ for the $i$-th saved checkpoint, the solved set, the tasks solved
by at least $k$ of the first
$i$ checkpoints, and the gap on domain $d$ are
\begin{align}
    Q_i^{(d)} &= \left\{x\in\mathcal{D}_{\rm val}^{(d)}: C_{T_i}(x)=1\right\},
    \label{eq:preliminary-capability}\\
    U_{\geq k}^{(d)}(i) &= \Bigl\{x\in\mathcal{D}_{\rm val}^{(d)}:
    \textstyle\sum_{j=1}^{i}\mathbf{1}\!\left[x\in Q_j^{(d)}\right]\geq k\Bigr\},
    \qquad
    G_i^{(d)} = \left|U_{\geq 1}^{(d)}(i)\setminus Q_i^{(d)}\right|,
    \label{eq:preliminary-gap}
\end{align}
and we write $U_{\geq k}^{(d)}=U_{\geq k}^{(d)}(n)$ for the whole run. So
$U_{\geq 1}^{(d)}(i)=\bigcup_{j\leq i}Q_j^{(d)}$ is the historical union through
checkpoint $i$, and $G_i^{(d)}$ counts the tasks some checkpoint up to that point
solves and $T_i$ does not. Raising $k$ drops the tasks only a single checkpoint
ever solved, so $U_{\geq 2}^{(d)}$ and $U_{\geq 3}^{(d)}$ separate repeatedly
solvable behavior from isolated successes.

The temporal capability gap persists across the evaluated checkpoints
(Figure~\ref{fig:preliminary-gap}). On DeepSearch it widens along training, and
in every domain the best checkpoint trails $U_{\geq 1}^{(d)}$ by the margin
annotated in Figure~\ref{fig:preliminary-gap}(b); per-domain curves are in
Appendix~\ref{app:gap-domains}. At step $500$, the final checkpoint solves
$2{,}808$ tasks, while $4{,}553$ tasks are solved by at least three checkpoints,
a difference of $62\%$.

Historical checkpoints solve complementary subsets of tasks. This motivates
selecting teachers by their coverage of the student's current failures.

\section{\method}
\label{sec:method}

\begin{figure}[t]
    \centering
    \includegraphics[width=\linewidth]{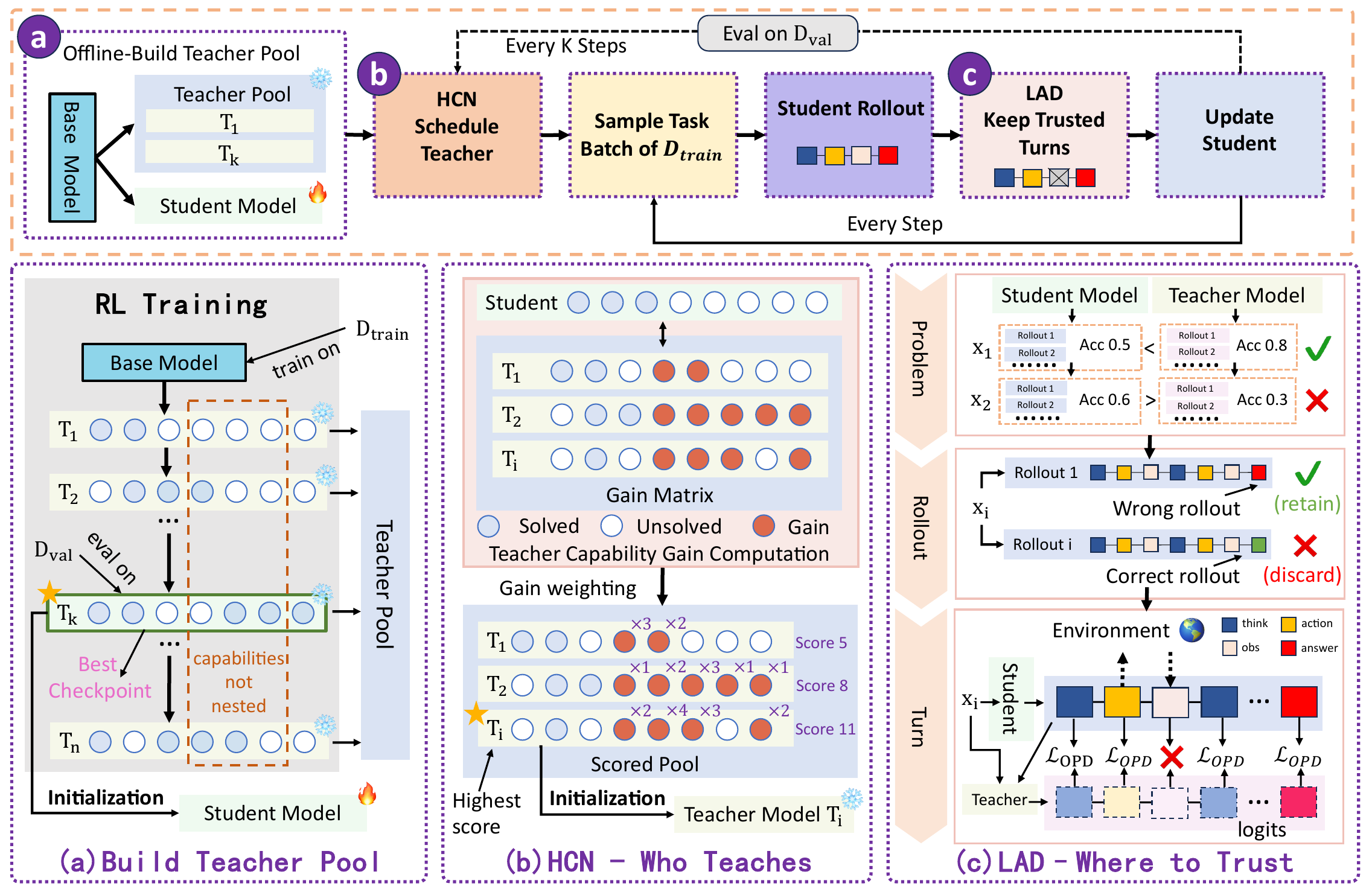}
    \caption{Overview of \method. \textbf{Top:} the consolidation loop.
    \textbf{(a)} Frozen checkpoints from one RL run form the teacher pool.
    \textbf{(b)} HCN selects teachers based on coverage of the student's unsolved tasks.
    \textbf{(c)} LAD filters supervision at the problem, rollout, and turn levels.}
    \label{fig:overview}
\end{figure}

The checkpoint-level complementarity observed in Section~\ref{sec:preliminary}
motivates \method.
The student is initialized from the best-validation RL checkpoint and trained with OPD.
Figure~\ref{fig:overview} illustrates the full pipeline: How to Build the Teacher Pool
~(a, Section~\ref{sec:teacher-pool}), teacher scheduling via
HCN~(b, Section~\ref{sec:hcn}), supervision filtering via
LAD~(c, Section~\ref{sec:lad}), and the consolidation loop that assembles the three
(Section~\ref{sec:consolidation-loop}).

\subsection{Build Teacher Pool}
\label{sec:teacher-pool}

Let $\{T_i\}_{i=1}^{n}$ be the checkpoints of one RL run in training-step order,
instantiated with GRPO, in which every rollout is scored by the verifiable outcome
indicator of Eq.~\ref{eq:preliminary-correctness}. We write $r(x,\hat y)\in\{0,1\}$
for that indicator applied to an individual rollout answer $\hat y$ rather than to a
policy's decoded answer, so that $C_{\pi}(x)=r\!\left(x,\hat y_{\pi}(x)\right)$.

Unless otherwise specified, we save checkpoints every 10 RL steps and freeze all saved checkpoints from
the complete run into the teacher pool. The student is initialized from the best-validation RL
checkpoint. HCN uses $\mathcal{D}_{\rm val}$ to select teachers according to
the problems they solve that the current student misses
(Eq.~\ref{eq:teacher-routing}).

\subsection{HCN: Which Teacher to Distill From}
\label{sec:hcn}

With multiple frozen candidates in the pool, the order in which they teach the
student matters. We propose \textbf{Highest Coverage Next} (HCN), a scheduling
policy for activating them.

\paragraph{Gain matrix.}
Evaluating a policy on $\mathcal{D}_{\rm val}$ splits it into the problems that policy
solves and those it does not; a problem that a candidate solves and the student does
not is a \textbf{gain} of that candidate. Consolidation proceeds in scheduling steps
$t$, one every $K$ training steps. Let $\student$ denote the student policy at
scheduling stage $t$. We record each candidate's solved set once and
refresh the student's at each of them,
\begin{align}
    Q_i &= \left\{x\in\mathcal{D}_{\rm val}: C_{T_i}(x)=1\right\},\qquad i\in[n],
    \label{eq:teacher-capability}\\
    P_t &= \left\{x\in\mathcal{D}_{\rm val}: C_{\student}(x)=1\right\},
    \label{eq:student-capability}
\end{align}
where HCN uses these sets to compute teacher scheduling scores. The gains of
candidate $i$ at scheduling step $t$ are then
\begin{equation}
    \Delta_{i,t}=Q_i\setminus P_t,
    \label{eq:marginal-coverage}
\end{equation}
and stacking $\Delta_{1,t},\dots,\Delta_{n,t}$ as rows gives the gain matrix of
Figure~\ref{fig:overview}(b).

\paragraph{Scored pool.}
Each problem $x$ carries a priority weight $w_t(x)$, initialized to $w_0(x)=1$.
The weighted coverage of candidate $i$ is
\begin{equation}
    g_{i,t}=\sum_{x\in\Delta_{i,t}} w_t(x).
    \label{eq:weighted-coverage}
\end{equation}
After each scheduling step the weights are updated by
\begin{equation}
    w_{t+1}(x)=
    \begin{cases}
        0, & x\in P_{t+1},\\
        1, & x\in Q_{i_t^\ast}\setminus P_{t+1},\\
        \min\{w_t(x)+\eta,\,w_{\max}\}, & \text{otherwise},
    \end{cases}
    \label{eq:routing-weight-update}
\end{equation}
with increment $\eta>0$ and cap $w_{\max}$.
Problems solved by the updated student receive zero weight. Unsolved problems
covered by the current teacher return to base priority. The update increases the
priority of problems that remain unsolved and are not covered by the current
teacher, up to $w_{\max}$.

\paragraph{Scheduling teacher.}
HCN activates the candidate with the highest weighted coverage among those that
still cover something the student misses,
\begin{equation}
    T_t^\ast = T_{i_t^\ast},
    \qquad
    i_t^\ast = \arg\max_{i\in\mathcal{C}_t} g_{i,t},
    \qquad
    \mathcal{C}_t=\left\{i\in[n]: g_{i,t}>0\right\},
    \label{eq:teacher-routing}
\end{equation}
with ties broken toward the earlier checkpoint. The activated teacher serves as the
supervisor for the next $K$ training steps, after which the student is re-evaluated on
$\mathcal{D}_{\rm val}$, $P_{t+1}$ is refreshed, and the process repeats.
Consolidation stops when $\mathcal{C}_t=\emptyset$.

\subsection{LAD: Where to Trust the Activated Teacher}
\label{sec:lad}

A teacher selected by HCN covers problems the student currently misses, but its
advantage may vary across training problems. We propose
\textbf{Locality-Aware Distillation} (LAD), which selects supervision through
filters at three levels shown top to bottom in
Figure~\ref{fig:overview}(c).

\paragraph{Problem level.} For a problem $x\in\mathcal{D}_{\rm train}$ the student and
the activated teacher each draw $G$ rollouts, both scored by the outcome reward
$r$ that the RL run already computes:
\begin{equation}
    s(x)=\frac{1}{G}\sum_{j=1}^{G} r\!\left(x,\hat y_S^{(j)}\right),
    \qquad
    q_t(x)=\frac{1}{G}\sum_{j=1}^{G} r\!\left(x,\hat y_{T_t^\ast}^{(j)}\right).
    \label{eq:pass-rates}
\end{equation}
We retain problems for which the teacher has a higher empirical success rate than
the student,
\begin{equation}
    \mathcal{X}_t=\left\{x : q_t(x)>s(x)\right\},
    \label{eq:problem-gate}
\end{equation}
which excludes problems where the teacher's empirical success rate does not exceed
the student's, including those for which both rates are zero. In the top block of
Figure~\ref{fig:overview}(c), $x_1$ is admitted and $x_2$ is rejected because the
student has a higher empirical success rate on it.

\paragraph{Rollout level.} Inside an admitted problem only the rollouts the student
failed carry a gap to close,
\begin{equation}
    \mathcal{F}(x)=\left\{ j\in[G] :
    r\!\left(x,\hat y_S^{(j)}\right)=0\right\},
    \qquad x\in\mathcal{X}_t.
    \label{eq:failure-mask}
\end{equation}
The teacher's $G$ rollouts estimate $q_t(x)$; distillation uses the student's
failed trajectories.
The middle block of Figure~\ref{fig:overview}(c) retains the student's wrong
rollout and discards the correct one.

\paragraph{Turn level.} The frozen teacher runs a single forward pass over each
trajectory $\tau$ retained by Eqs.~\ref{eq:problem-gate} and~\ref{eq:failure-mask}.
Because an agentic trajectory interleaves student turns with environment returns,
the loss is computed over the student-generated positions $\mathcal{G}(\tau)$,
comprising the think, action, and answer turns in the bottom block of
Figure~\ref{fig:overview}(c). Prompt tokens, observations, tool returns, and padding
are excluded. At such a position $\ell$ we truncate to the student's top-$k$ tokens $A_\ell^{(k)}$ and
renormalize both policies on that support,
$p^{(k)}_{\pi}(v\mid\tau_{<\ell})=\pi(v\mid\tau_{<\ell})\big/
\sum_{u\in A_\ell^{(k)}}\pi(u\mid\tau_{<\ell})$,
and minimize $D_{\mathrm{KL}}(p_S^{(k)}\|p_T^{(k)})$
\begin{equation}
    \mathcal{L}_{\mathrm{OPD}}
    = \frac{1}{\sum_{\tau}\bigl|\mathcal{G}(\tau)\bigr|}
      \sum_{\tau}\ \sum_{\ell\in\mathcal{G}(\tau)}
      D_{\mathrm{KL}}\!\left(
        p^{(k)}_{S,t}\!\left(\cdot\mid\tau_{<\ell}\right)\;\middle\|\;
        p^{(k)}_{T_t^\ast}\!\left(\cdot\mid\tau_{<\ell}\right)\right).
    \label{eq:opd-loss}
\end{equation}

\subsection{The Consolidation Loop: How Much to Draw per Update}
\label{sec:consolidation-loop}

The three components compose into the loop at the top of Figure~\ref{fig:overview}: HCN
activates a teacher, the student rolls out problems drawn from $\mathcal{D}_{\rm train}$,
LAD selects the samples worth learning from, and the student is updated on them. Since
LAD discards most of what is generated, we follow the oversampling of DAPO
\citep{yu2025dapo} and keep drawing batches of $B$ problems until the sequences retained
by Eqs.~\ref{eq:problem-gate} and~\ref{eq:failure-mask} fill one batch,
discarding the surplus, with a draw capped at $L$ batches. We use a fixed number of
retained sequences per update and retain the optimizer hyperparameters of the
unfiltered setting. Additional sampling and teacher evaluation incur inference
cost (Appendix~\ref{app:cost}).

\section{Experiments}
\label{sec:experiments}

\begin{table}[t]
\centering
\caption{DeepSearch exact match accuracy (\%) on the test sets
($\mathcal{D}_{\rm test}$) of seven benchmarks.
\textbf{Bold} marks the best score within each backbone block. Reference models
use the same retriever, four-turn limit, and scoring, without domain-specific
training.}
\label{tab:main-results}
\small
\setlength{\tabcolsep}{3pt}
\resizebox{\linewidth}{!}{%
\begin{tabular}{@{}l@{\hspace{9pt}}cccccccc@{}}
\toprule
& \multicolumn{2}{c}{In-domain} & \multicolumn{5}{c}{Out-of-domain} & \\
\cmidrule(lr){2-3} \cmidrule(lr){4-8}
Model & NQ & HotpotQA & PopQA & TriviaQA & 2Wiki & MuSiQue & Bamboogle & Avg. \\
\midrule
\multicolumn{9}{@{}l}{\textbf{Proprietary models}} \\
GPT-5.6~\citep{openai2026gpt56}            &  7.7 & 22.2 & 32.2 & 56.5 & 17.4 & 11.2 & 30.4 & 25.4 \\
Claude Sonnet 5~\citep{anthropic2026claudesonnet5}  & 23.0 & 16.8 & 22.6 & 64.8 &  4.5 &  5.9 & 40.8 & 25.5 \\
Grok-4.6~\citep{xai2026grok46}           & 20.7 & 33.4 & 27.3 & 59.8 & 25.6 & 12.3 & 46.4 & 32.2 \\
\midrule
\multicolumn{9}{@{}l}{\textbf{Open-source models, ${>}100$B parameters}} \\
DeepSeek-V4.1~\citep{xu2026deepseek}          & 31.5 & 42.6 & 40.5 & 72.9 & 30.7 & 15.9 & 64.8 & 42.7 \\
GLM-5.3-Flash~\citep{glm5team2026glm5vibecodingagentic}          & 29.1 & 46.7 & 43.9 & 70.5 & 36.6 & 23.3 & 64.8 & 45.0 \\
Qwen3.5-122B-A10B~\citep{qwen3.5}      & 34.4 & 42.1 & 42.4 & 72.0 & 36.2 & 20.4 & 59.2 & 43.8 \\
\addlinespace[2pt]
\multicolumn{9}{@{}l}{\textbf{Open-source models, $20$--$100$B parameters}} \\
Llama-3.1-70B-Instruct~\citep{grattafiori2024llama} & 32.3 & 37.7 & 37.5 & 63.5 & 30.1 & 16.1 & 48.0 & 37.9 \\
Qwen3.6-35B-A3B~\citep{qwen36_35b_a3b}        & 34.8 & 38.6 & 41.0 & 68.1 & 30.0 & 17.5 & 56.0 & 40.9 \\
Qwen3.5-35B-A3B~\citep{qwen3.5}        & 32.1 & 39.7 & 38.9 & 70.3 & 30.0 & 18.8 & 53.6 & 40.5 \\
Qwen3.8-27B~\citep{qwen38}            & 33.2 & 41.0 & 36.2 & 56.7 & 26.5 & 15.0 & 48.0 & 36.0 \\
Qwen3.6-27B~\citep{qwen36_35b_a3b}            & 32.7 & 38.2 & 38.6 & 65.5 & 29.0 & 15.9 & 52.0 & 38.9 \\
Qwen3.5-27B~\citep{qwen3.5}            & 31.1 & 39.5 & 36.1 & 67.8 & 31.9 & 17.5 & 52.8 & 39.5 \\
\addlinespace[2pt]
\multicolumn{9}{@{}l}{\textbf{Open-source models, ${\le}20$B parameters}} \\
Qwen3.5-9B~\citep{qwen3.5}             & 23.7 & 28.2 & 28.3 & 56.9 & 28.5 & 10.7 & 38.4 & 30.7 \\
Llama-3.1-8B-Instruct~\citep{grattafiori2024llama}  & 28.7 & 28.3 & 33.5 & 52.3 & 21.0 & 10.7 & 43.2 & 30.4 \\
\midrule
\multicolumn{9}{@{}l}{\textbf{Backbone: Qwen2.5-3B-Instruct}~\citep{qwen2.5}} \\
\quad Base model    & 27.9 & 21.2 & 33.3 & 49.6 & 21.0 &  6.6 & 22.4 & 26.0 \\
\quad GRPO          & 42.1 & 34.8 & 39.2 & 55.5 & \textbf{35.8} & 13.9 & 28.0 & 35.6 \\
\quad OPSD         & 24.8 & 21.0 & 31.5 & 45.9 & 23.3 &  4.5 & 24.8 & 25.1 \\
\quad \method & \textbf{45.4} & \textbf{37.9} & \textbf{43.2} & \textbf{60.8} & 34.3 & \textbf{14.6} & \textbf{32.8} & \textbf{38.4} \\
\midrule
\multicolumn{9}{@{}l}{\textbf{Backbone: Qwen2.5-7B-Instruct}~\citep{qwen2.5}} \\
\quad Base model    & 26.0 & 28.3 & 26.8 & 53.0 & 25.9 & 12.4 & 35.2 & 29.6 \\
\quad GRPO          & 43.2 & 39.2 & 41.7 & 60.9 & 29.9 & 16.3 & 40.8 & 38.8 \\
\quad OPSD         & 27.4 & 29.4 & 27.3 & 52.8 & 26.3 & 12.9 & 32.8 & 29.9 \\
\quad \method & \textbf{50.4} & \textbf{46.5} & \textbf{47.8} & \textbf{66.5} & \textbf{41.1} & \textbf{22.5} & \textbf{48.8} & \textbf{46.2} \\
\midrule
\multicolumn{9}{@{}l}{\textbf{Backbone: Qwen3-4B-Instruct-2507}~\citep{qwen3technicalreport}} \\
% Base model and GRPO are the Qwen3-4B-Instruct-2507 rows of the 20260922
% comparison eval; GRPO is step 585, the checkpoint \method initialises from.
\quad Base model    & 25.7 & 24.1 & 28.6 & 49.0 & 25.1 &  6.2 & 37.6 & 28.0 \\
\quad GRPO          & 50.4 & 49.2 & 48.6 & 66.4 & 45.3 & 22.6 & 52.0 & 47.8 \\
\quad OPSD          & 30.4 & 30.9 & 34.8 & 52.3 & 17.3 &  9.4 & 25.6 & 28.7 \\
\quad \method  & \textbf{55.6} & \textbf{52.2} & \textbf{50.8} & \textbf{69.1} & \textbf{46.1} & \textbf{24.4} & \textbf{54.8} & \textbf{50.4} \\
\bottomrule
\end{tabular}%
}
\end{table}

To verify the stability and effectiveness of \method, we conduct extensive
experiments across a wide range of scenarios and environments. 

\begin{figure}[t]
    \centering
    % Match the panel aspect ratios; subcaptions are typeset by LaTeX.
    \begin{minipage}[b]{0.48\linewidth}
        \centering
        \includegraphics[width=\linewidth]{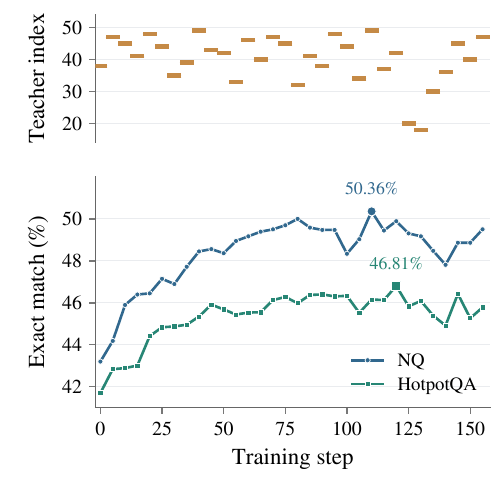}\\[4pt]
        {\small (a) Teacher selection and QA performance.\\Base: Qwen2.5-7B-Instruct.}
    \end{minipage}%
    \hfill
    \begin{minipage}[b]{0.48\linewidth}
        \centering
        \includegraphics[width=\linewidth]{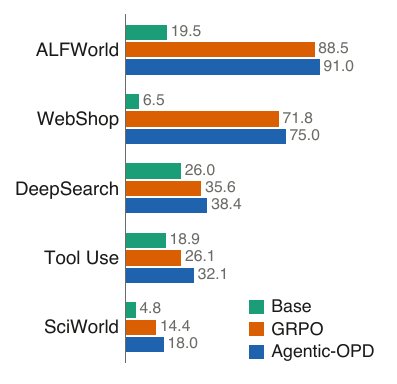}\\[4pt]
        {\small (b) Five-domain overview.\\Base: Qwen2.5-3B-Instruct.}
    \end{minipage}
    \caption{\textbf{(a)} Teacher indices and NQ/HotpotQA exact match during
    consolidation; highlighted points mark peak accuracy.
    \textbf{(b)} Domain scores (\%): mean exact match for DeepSearch, mean pass@1
    for Tool-use, success rate for ALFWorld, mean of Score and Success for
    WebShop, and Overall score for SciWorld. Base denotes the instruction-tuned model.}
    \label{fig:domain-overview}
\end{figure}

\subsection{Experimental Setup}
\label{sec:exp-settings}

\textbf{Tasks and Metrics.} We evaluate \method on DeepSearch, Tool-use,
WebShop, SciWorld, and ALFWorld. DeepSearch uses a Wikipedia retriever and is
scored by exact match on seven question answering benchmarks. Tool-use provides
a Python interpreter and is evaluated by pass@1 on AIME 2024, AIME 2025,
MATH-500, AMC23, and OlympiadBench. We report task score and success rate for
WebShop, Overall score for SciWorld, and success rate for ALFWorld.
Appendix~\ref{app:env-settings} lists the benchmarks, interaction limits, and
evaluation settings.

\textbf{Models and Baselines.} We use Qwen2.5-3B-Instruct,
Qwen2.5-7B-Instruct, and Qwen3-4B-Instruct for DeepSearch. The other four domains
use Qwen2.5-3B-Instruct. All ablation and model-merging experiments use
Qwen2.5-3B-Instruct.
We compare with the instruction-tuned base model and GRPO.
The DeepSearch comparison also includes OPSD~\citep{zhao2026self}, which uses
privileged information to construct teacher supervision.
Table~\ref{tab:main-results} additionally reports proprietary and open-source
reference models evaluated without domain-specific training.

\textbf{Training and Evaluation.} RL and OPD use the same training data.
The student is initialized from the best-validation checkpoint of each RL run.
For DeepSearch, we use 90\% of the NQ and HotpotQA training sets for training
and reserve the remaining 10\% as $\mathcal{D}_{\rm val}$ for teacher selection
and the preliminary capability analysis (Section~\ref{sec:preliminary}).
All results in Table~\ref{tab:main-results} are measured on the corresponding
benchmark test sets $\mathcal{D}_{\rm test}$. RL training follows Search-R1~\citep{jin2025search} for
DeepSearch, ReTool~\citep{feng2026retool} for Tool-use, and AgentGym-RL~\citep{xi2025agentgym} for
WebShop, SciWorld, and ALFWorld. Unless otherwise specified, we save a checkpoint every 10 RL steps and
include all saved checkpoints from the full RL run in the teacher pool.
OPD uses the same rollout sampling settings as RL and minimizes $D_{\mathrm{KL}}(p_S^{(k)}\|p_T^{(k)})$
between the student and the selected teacher, as defined in Eq.~\ref{eq:opd-loss}.
Evaluation follows the inference settings of the corresponding RL method.
All training runs use one node with 8$\times$A800 80GB GPUs.

\subsection{Effectiveness and Generality Across Agentic Domains}
\label{sec:exp-effectiveness}

\method improves mean DeepSearch accuracy over GRPO by 2.8, 7.4, and 2.6
percentage points on Qwen2.5-3B, Qwen2.5-7B, and Qwen3-4B, respectively
(Table~\ref{tab:main-results}), with gains also extending to out-of-domain
benchmarks. Historical checkpoints thus retain useful capabilities that the
student can acquire through distillation after RL training.

Table~\ref{tab:main-results} also lists reference models evaluated with the
same retriever, four-turn limit, and exact-match scoring, without training on
this protocol. Exact match can penalize answers that are correct but not given
as a short span, so these scores reflect fit to the protocol rather than general
capability. We include them as context and do not rank \method against them.

OPSD stays within one point of the base model on all three backbones (25.1,
29.9, and 28.7 versus 26.0, 29.6, and 28.0) and well below GRPO, suggesting that
privileged information for a teacher using the same model as the student is
insufficient for effective distillation in these agentic tasks. \method instead schedules historical
teachers according to the student's current capability gaps and restricts
distillation to supervision where the teacher has a local advantage.
Figure~\ref{fig:domain-overview}(a) shows that teacher selection changes as the
student develops and accuracy improves during consolidation, supporting
selection based on complementary capabilities.

Across 24 benchmarks and task subsets spanning five agentic domains, \method improves aggregate
performance over GRPO in every domain (Figure~\ref{fig:domain-overview}(b)).
The gains extend beyond retrieval-based question answering to code execution
and multi-step environment interaction, showing that historical checkpoint
reuse can improve a single policy across different agentic tasks.

\subsection{Capability Recovery and Teacher Complementarity}
\label{sec:exp-recovery}

We examine the Qwen3-4B consolidation run on 2,000 routing-validation
questions, with 1,000 each from NQ and HotpotQA. This analysis uses the
100 historical teachers saved every five RL steps through step 500, rather
than the default ten-step interval, and the
best observed OPD validation checkpoint, update 20. Since this checkpoint is
selected on the same validation set, the analysis describes consolidation
behavior rather than independent test-set performance.

\begin{figure}[t]
    \centering
    \includegraphics[width=\linewidth]{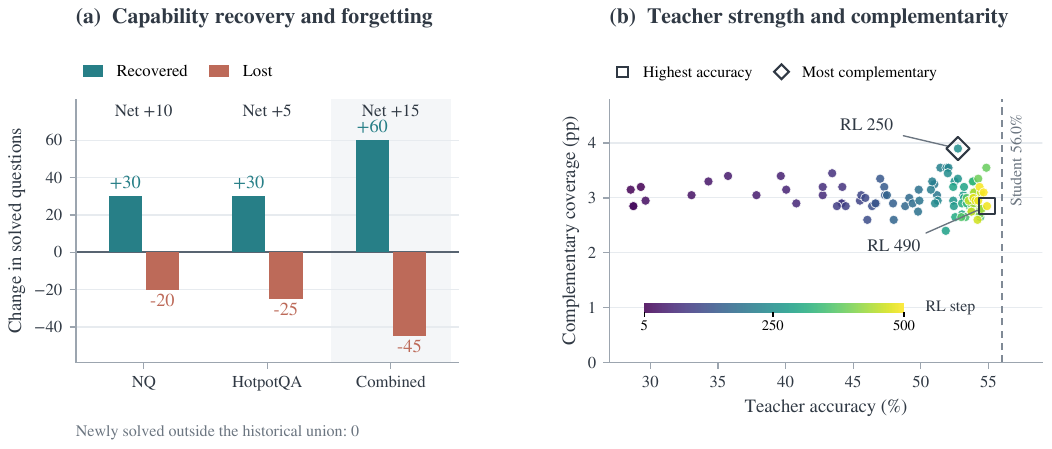}
    \caption{Capability recovery and teacher complementarity on the Qwen3-4B
    routing-validation set. \textbf{(a)} Question-level changes after OPD
    update 20. \textbf{(b)} Teacher accuracy versus unweighted coverage
    $100|Q_i\setminus P_{20}|/2{,}000$. Color indicates RL step; the dashed
    line marks student accuracy.}
    \label{fig:recovery-complementarity}
\end{figure}

\paragraph{Recovery and retention.}
Figure~\ref{fig:recovery-complementarity}(a) separates gains from losses at the
question level. The student recovers 60 questions from the historical union
while losing 45 previously solved questions, for a net gain of 15 questions
(+0.75 percentage points, from 55.25\% to 56.0\%). All newly solved questions
were solved by at least one historical teacher. Thus the observed improvement
coincides with recovery of historical coverage, although substantial losses
remain even at the best observed validation checkpoint.

\paragraph{Strength and complementarity.}
Figure~\ref{fig:recovery-complementarity}(b) compares each teacher's overall
accuracy with its coverage of the best student's remaining failures.
The highest-accuracy teacher, RL step 490, scores 54.9\% and covers 57 such
questions (2.85 points). The most complementary teacher, step 250, has lower
accuracy at 52.75\% but covers 78 questions (3.9 points). Overall teacher
accuracy therefore does not determine its usefulness for the student's
remaining gaps, supporting the motivation for coverage-based scheduling.

\subsection{Checkpoint Consolidation Analysis}
\label{sec:exp-ablation}
\label{sec:exp-merging}

\begin{wraptable}{r}{0.42\linewidth}
\centering
\caption{Checkpoint reuse and ablations on DeepSearch with Qwen2.5-3B-Instruct: exact match (\%).}
\label{tab:merging}
\small
\setlength{\tabcolsep}{4pt}
\begin{tabular}{@{}lcc@{}}
\toprule
Method & NQ & HotpotQA \\
\midrule
GRPO (initialization) & 42.1 & 34.8 \\
\midrule
\multicolumn{3}{@{}l}{\textit{Training-free merging}} \\
Uniform soup       & 42.5 & 35.1 \\
Greedy soup        & 43.4 & 35.9 \\
TIES               & 43.0 & 35.6 \\
DARE-TIES          & 43.7 & 36.2 \\
EMA (last $k$)     & 42.8 & 35.5 \\
\addlinespace[2pt]
\multicolumn{3}{@{}l}{\textit{Merging plus RL}} \\
GTR-Turbo          & 45.0 & 37.4 \\
\addlinespace[2pt]
\multicolumn{3}{@{}l}{\textit{Inference ensemble}} \\
Temporal Sampling  & 44.6 & 37.1 \\
\midrule
\method            & \textbf{45.4} & \textbf{37.9} \\
\quad w/o HCN      & 43.6 & 36.1 \\
\quad w/o LAD      & 42.7 & 35.3 \\
\bottomrule
\end{tabular}
\end{wraptable}

Table~\ref{tab:merging} compares checkpoint reuse methods and ablations
on NQ and HotpotQA with Qwen2.5-3B-Instruct.

\paragraph{Teacher scheduling and filtering.}
Replacing HCN with uniform teacher sampling at the same scheduling frequency
reduces accuracy by 1.8 percentage points on both datasets. Removing LAD at
the problem, rollout, and turn levels causes larger drops of 2.7 and 2.6
points on NQ and HotpotQA, respectively. Both variants still improve over
GRPO, but lose much of the gain from consolidation. These results support
selecting teachers for the student's remaining gaps and filtering their
supervision according to local advantage.

\paragraph{Alternative reuse methods.}
We compare model soups~\citep{wortsman2022model}, TIES and
DARE-TIES~\citep{yadav2023ties,yu2024language}, checkpoint
EMA~\citep{sanyal2023early}, GTR-Turbo's alternating merging and RL
updates~\citep{wei2026gtr}, and Temporal Sampling~\citep{li2026temporal}.
All methods use the same GRPO run; greedy soup has the same access to
validation data as \method.
\method achieves the highest accuracy on both datasets, improving over GRPO
by 3.3 points on NQ and 3.1 points on HotpotQA. DARE-TIES is the strongest
merging method without further training, but trails \method by 1.7 points
on each dataset. GTR-Turbo narrows the gap to 0.4 and 0.5 points with
additional RL updates. \method also exceeds Temporal Sampling by 0.8 points
on both datasets while using a single policy at inference. Thus, distilling
selected historical teachers provides greater gains than the other reuse
methods evaluated here.

\section{Conclusion}

The best checkpoint from an agentic RL run does not retain everything learned
during training. Across five domains, historical checkpoints solve many tasks
that the selected checkpoint misses. We use these checkpoints as frozen
teachers to recover some of this lost coverage in a single student. HCN selects
teachers according to the student's remaining failures, while LAD restricts
distillation to problems where the teacher performs better and to failed
student trajectories. The resulting policy improves aggregate performance over
GRPO in all five domains, with gains across three backbones on DeepSearch.
The ablations show why both decisions matter: removing either teacher
scheduling or supervision filtering reduces the gains over the student's
initialization on the evaluated NQ and HotpotQA benchmarks.
These results support retaining historical checkpoints as a source of useful
supervision after RL training. A checkpoint can still teach the student even
when its overall accuracy is lower. Recovery remains incomplete, however,
and some tasks show regressions. Consolidation also requires additional
sampling and evaluation, although deployment uses only the student. Reducing
this cost and measuring how much of the historical task coverage can be
reliably recovered are useful next steps.

\section{Related Work}
\label{sec:related}

\subsection{Agentic RL and On-Policy Distillation}

Verifiable-reward RL trains language models for multi-turn search, tool use, and
other interactive tasks. Representative systems include Search-R1, s3,
Agentic-R, ReTool, and AgentGym-RL
\citep{jin2025search,jiang2025s3,liu2026agentic,feng2026retool,xi2025agentgym},
with GRPO and its variants providing common optimization frameworks
\citep{shao2024deepseekmath,yu2025dapo}. These runs produce a sequence of
checkpoints; \method treats that sequence as a source of complementary
capabilities rather than using only one checkpoint.

Sequence-level distillation trains on teacher-generated trajectories, whereas
on-policy distillation evaluates the student's own prefixes with teacher token
distributions. GKD and MiniLLM established this paradigm, and recent work
extends it to multi-turn agents through step-wise or turn-aware objectives
\citep{agarwal2024policy,gu2024minillm,zhong2026sod,zhou2026turnopd}.
Multi-teacher methods study capability integration and conflicts among
complementary teachers, including counteraction-aware and teacher-debate-based
approaches
\citep{ma2026mopd,chen2026counteraction,wang2026mad,men2026physics}.
Most use parallel or independently developed teachers. \method instead uses
historical checkpoints as frozen temporal teachers: HCN selects them by the
student's remaining capability gaps, and LAD restricts supervision to problems
where the teacher has a local advantage and to failed student trajectories.

\subsection{Historical Checkpoint Reuse and Model Merging}

Model merging combines compatible models or checkpoints to aggregate
capabilities. Model Soups averages weights from multiple fine-tuned models,
while Task Arithmetic, Fisher Merging, TIES, and DARE improve parameter-space
composition through task directions, parameter importance, interference
resolution, and sparsification, respectively
\citep{wortsman2022model,ilharco2022editing,matena2022merging,yadav2023ties,
yu2024language}. More recent work extends merging to iterative model updating
through activation-guided rotation-aware merging
\citep{yao2026merging}.

A related line of work reuses historical checkpoints from a single training
trajectory. Checkpoint averaging can improve optimization stability and
generalization \citep{sanyal2023early,li2026model}, while Temporal Sampling
recovers forgotten reasoning through outputs from multiple SFT or RL
checkpoints \citep{li2026temporal}. GTR-Turbo merges historical checkpoints
into a unified teacher for subsequent RL
\citep{wei2026gtr}, and AgentPatch studies capability degradation after merging
specialized agentic MLLMs and performs training-free capability repair
\citep{shao2026agentpatch}. These approaches either reuse checkpoint outputs or
collapse checkpoints into a single model. \method keeps them as separate frozen
teachers and routes supervision according to the student's capability gaps and
each teacher's local advantage, consolidating the recovered behavior into one
deployable student.

\clearpage
\bibliography{iclr2027_conference}
\bibliographystyle{iclr2027_conference}

\clearpage
\appendix
\section{Appendix}

\subsection{Detailed Environment Settings}
\label{app:env-settings}

RL training follows Search-R1~\citep{jin2025search} for DeepSearch,
ReTool~\citep{feng2026retool} for Tool-use, and AgentGym-RL for WebShop,
SciWorld, and ALFWorld. OPD uses the same training data and rollout sampling
settings as RL. Unless otherwise specified, the teacher pool contains all checkpoints saved at intervals
of 10 RL steps. OPD minimizes $D_{\mathrm{KL}}(p_S^{(k)}\|p_T^{(k)})$ in Eq.~\ref{eq:opd-loss}.

\begin{table}[h]
\centering
\caption{Benchmarks, evaluation metrics, and maximum interaction turns per episode.}
\label{tab:domains}
\small
\setlength{\tabcolsep}{4pt}
\begin{tabular}{@{}l >{\raggedright\arraybackslash}p{0.42\linewidth} l c@{}}
\toprule
Domain & Benchmarks / tasks & Metric & Turns \\
\midrule
DeepSearch
  & NQ~\citep{kwiatkowski2019natural}, HotpotQA~\citep{yang2018hotpotqa}
    (in-domain); PopQA~\citep{mallen2023not},
    TriviaQA~\citep{joshi2017triviaqa}, 2Wiki~\citep{ho2020constructing},
    MuSiQue~\citep{trivedi2022musique}, Bamboogle~\citep{press2023measuring}
    (out-of-domain)
  & Exact match & 4 \\
\addlinespace[3pt]
Tool-use
  & AIME 2024, AIME 2025, MATH-500~\citep{lightman2024let}, AMC23,
    OlympiadBench~\citep{he2024olympiadbench}
  & pass@1 & 16 \\
\addlinespace[3pt]
WebShop~\citep{yao2022webshop}
  & Goal-directed shopping on simulated e-commerce pages
  & Score, success rate & 20 \\
\addlinespace[3pt]
SciWorld~\citep{wang2022scienceworld}
  & Measure, Test-Cond., Find, Chem-Mix, Lifespan
  & Overall score & 15 \\
\addlinespace[3pt]
ALFWorld~\citep{shridhar2020alfworld}
  & Embodied household instruction following
  & Success rate & 50 \\
\bottomrule
\end{tabular}
\end{table}

\paragraph{DeepSearch.} The agent retrieves passages from a Wikipedia index
through a search tool. We use 90\% of the NQ and HotpotQA training sets for
both RL and OPD and reserve the remaining 10\% for teacher selection.
Evaluation covers the seven benchmarks in Table~\ref{tab:domains} and follows
the inference protocol of Search-R1~\citep{jin2025search}. Each episode allows
up to 4 interactions.

\paragraph{Tool-use.} The agent executes Python code in a sandboxed
interpreter during reasoning. Training and evaluation follow
ReTool~\citep{feng2026retool}, with at most 16 interaction turns per episode.
For evaluation, we use temperature 1.0 and top-$p$ 0.7, repeat evaluation
32 times, and report the average accuracy as an estimate of pass@1.

\paragraph{WebShop.} The agent searches for products and navigates simulated
shopping pages to complete a purchase instruction. Each episode allows up to
20 interaction turns. We report both task score and success rate.

\paragraph{SciWorld.} The agent interacts with a text-based scientific
environment. We report scores for Measure, Test-Cond., Find, Chem-Mix, and
Lifespan, together with the Overall score. Each episode allows up to 15
interaction turns.

\paragraph{ALFWorld.} The agent follows household instructions in a text-based
environment. We evaluate Pick, Look, Clean, Heat, Cool, and Pick2, with at most
50 interaction turns per episode. We report success rates for each task type
and for the full evaluation set.

For WebShop, SciWorld, and ALFWorld, RL training follows AgentGym-RL.
Evaluation uses temperature 0.7, top-$p$ 0.8, and top-$k$ 20. Results are
averaged over three random seeds.

\subsection{Temporal Capability Gap Across Domains}
\label{app:gap-domains}

\begin{figure}[h]
    \centering
    \includegraphics[width=\linewidth]{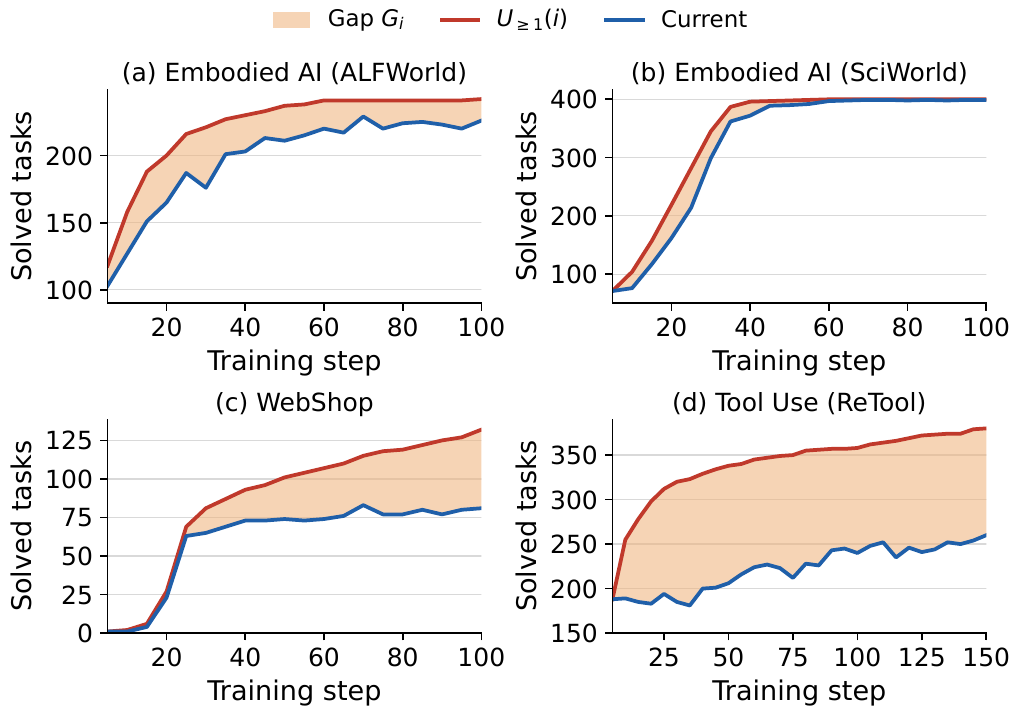}
    \caption{Validation tasks solved during RL in four domains: current checkpoint
    (blue) and historical union $U_{\geq 1}^{(d)}(i)$ (red).
    Shading marks the gap $G_i^{(d)}$.}
    \label{fig:appendix-gap}
\end{figure}

Figure~\ref{fig:appendix-gap} repeats the measurement of
Section~\ref{sec:preliminary} on four further environments. For each domain we
train with GRPO under result-based rewards, evaluate every saved checkpoint on
the domain's validation set, and plot the count solved by the current
checkpoint, $|Q_i^{(d)}|$, against the historical union up to that step,
$U_{\geq 1}^{(d)}(i)$; the shaded band is the gap $G_i^{(d)}$ of
Eq.~\ref{eq:preliminary-gap}. Step 0 is the untrained checkpoint and is
excluded, and the vertical axes are truncated below the lowest observed count.

The gap is present in every domain and does not close with training. It is
widest on ReTool and WebShop, where the current checkpoint plateaus while
$U_{\geq 1}^{(d)}(i)$ keeps growing, and narrowest on SciWorld, where both
curves saturate near the size of the validation set and leave little room for
the union to extend. This matches the DeepSearch observation: even within a
single run, the latest checkpoint is not a superset of what earlier checkpoints
could do, so the temporal pool retains complementary and recoverable behaviors.

Because SciWorld has a relatively small validation set, its point estimates
have low statistical power and differences between checkpoints may fail to
reach conventional significance thresholds even when their trajectories are
visibly different. We therefore treat the SciWorld curves as descriptive
evidence rather than as a standalone significance claim. For checkpoint
comparisons, we use paired resampling over the same task instances (with a
95\% bootstrap confidence interval) and aggregate estimates across the three
random seeds; we do not interpret a non-significant SciWorld comparison as
evidence of equivalence. The qualitative conclusion is retained only when the
direction of the gap is consistent across seeds and is supported by the
per-task trajectories.

\begin{table}[h]
\centering
\caption{Tool-use pass@1 accuracy (\%). Mean is the average across five benchmarks.}
\label{tab:app-tooluse}
% Seven columns with "OlympiadBench" as a header overruns \linewidth by ~5pt at
% the default 6pt \tabcolsep; 5pt buys 14pt back, which is the smallest change
% that clears it. Scoped to this table environment.
\setlength{\tabcolsep}{5pt}
\begin{tabular}{lcccccc}
\toprule
Method & AIME 2024 & AIME 2025 & MATH-500 & AMC23 & OlympiadBench & Mean \\
\midrule
Base   & 3.0  & 3.1   & 43.0 & 27.6  & 17.9  & 18.9 \\
GRPO   & 7.9  & 12.3  & 49.5 & 36.3  & 24.8  & 26.1 \\
\method & \textbf{15.1} & \textbf{15.5} & \textbf{55.3} & \textbf{44.8} & \textbf{29.9} & \textbf{32.1} \\
\bottomrule
\end{tabular}
\end{table}

\begin{table}[h]
\centering
\caption{SciWorld scores by task subset and the aggregate Overall score.}
\label{tab:app-sciworld}
\begin{tabular}{lcccccc}
\toprule
Method & Measure & Test-Cond. & Find & Chem-Mix & Lifespan & Overall \\
\midrule
Base   & 1.1  & 4.2  & 8.4  & 4.7  & 10.6 & 4.8 \\
GRPO   & 7.6  & 9.3  & 22.9 & 15.4 & 41.1 & 14.4 \\
\method & \textbf{14.8} & \textbf{10.0} & \textbf{25.0} & \textbf{26.6} & \textbf{50.0} & \textbf{18.0} \\
\bottomrule
\end{tabular}
\end{table}

\begin{table}[h]
\centering
\caption{WebShop task score and success rate (\%). Mean averages the two metrics.}
\label{tab:app-webshop}
\begin{tabular}{lccc}
\toprule
Method & Score & Success & Mean \\
\midrule
Base   & 10.4 & 2.5  & 6.5 \\
GRPO   & 80.6 & 63.0 & 71.8 \\
\method & \textbf{84.0} & \textbf{66.0} & \textbf{75.0} \\
\bottomrule
\end{tabular}
\end{table}

\begin{table}[!t]
\vspace{-16pt}
\centering
\caption{ALFWorld success rate (\%) by task type. All denotes the full evaluation set.}
\label{tab:app-alfworld}
\begin{tabular}{lccccccc}
\toprule
Method & Pick & Look & Clean & Heat & Cool & Pick2 & All \\
\midrule
Base   & 45.7 & 5.3 & 10.8 & 8.0 & 10.7 & 17.8 & 19.5 \\
GRPO   & 97.8 & 73.7 & \textbf{94.6} & \textbf{96.0} & 82.1 & 80.0 & 88.5 \\
\method & \textbf{100.0} & \textbf{78.9} & \textbf{94.6} & 92.0 & \textbf{85.7} & \textbf{86.7} & \textbf{91.0} \\
\bottomrule
\end{tabular}
\end{table}

\subsection{Cost of the Consolidation Loop}
\label{app:cost}

The consolidation loop of Section~\ref{sec:consolidation-loop} targets a fixed
number of retained sequences per update and retains the optimizer hyperparameters
of the unfiltered OPD setting. For a target batch of $B_{\rm eff}=BG$ sequences,
if a fraction $\rho$ of the student's rollouts survives the filters of
Eq.~\ref{eq:problem-gate}--\ref{eq:failure-mask}, filling the batch requires
approximately $B/\rho$ sampled problems and $2BG/\rho$ rollouts, split evenly
between the student and the activated teacher. The teacher also performs a
forward pass over each retained student trajectory to provide token-level
supervision. These estimates apply when the batch is filled before the cap of
$L$ sampling batches; updates that reach the cap may use fewer than
$B_{\rm eff}$ retained sequences. Additional sampling, scoring, and teacher
evaluation incur inference cost. HCN also requires evaluating the frozen
teacher pool on the validation set and periodically re-evaluating the student
for teacher scheduling.

\subsection{Per-Benchmark Results}
\label{app:full-results}

Figure~\ref{fig:domain-overview}(b) reports one aggregate score per domain:
mean exact match across the seven DeepSearch benchmarks, mean pass@1 across the
five Tool-use benchmarks, full-evaluation-set success rate (\emph{All}) for
ALFWorld, the mean of Score and Success for WebShop, and the environment's
\emph{Overall} score for SciWorld. DeepSearch per-benchmark results are
reported in Table~\ref{tab:main-results}; the tables below give the
per-benchmark and per-task results for the other four domains. All runs
reported here start from
Qwen2.5-3B-Instruct; ``Base'' denotes the instruction-tuned model before
domain-specific training.

\paragraph{Cross-benchmark analysis.} On the tool-use benchmarks, the mean
pass@1 increases from $26.1\%$ with GRPO to $32.1\%$ with \method. The method
improves all five benchmarks. The largest gains are on AMC23 (from $36.3\%$ to
$44.8\%$) and AIME 2024 (from $7.9\%$ to $15.1\%$), whereas the gain on
AIME 2025 is smallest ($12.3\%$ to $15.5\%$).

The interactive tasks show a similar pattern. On SciWorld, the Overall score
rises from $14.4$ to $18.0$, mainly due to gains on Chem-Mix and Lifespan;
Test-Cond. changes only slightly. On WebShop, Score increases from $80.6\%$
to $84.0\%$ and Success from $63.0\%$ to $66.0\%$, raising their mean from
$71.8\%$ to $75.0\%$.

ALFWorld also shows task-level variation: four task types improve, one is
unchanged, and Heat drops from $96.0\%$ to $92.0\%$. The \emph{All} success
rate rises from $88.5\%$ to $91.0\%$, with the largest task-level gains on Look
and Pick2. These results support
an overall improvement, while also showing that the effect depends on the
benchmark and task type.

\subsection{Full-Page Trajectory Cases}
\label{sec:case-study}

Each case is rendered as a single self-contained image with a task panel and
an Agentic-OPD trajectory panel. Reasoning is summarized for readability;
retrieved passages are excerpted, with ``[...]'' marking omissions. Green
highlights supporting evidence or successful actions, while red highlights
errors. The two DeepSearch cases are validation rollouts collected at OPD
update 20, rather than trajectories from an optimizer batch. They illustrate
an entity-disambiguation failure and a successful two-hop search, respectively.

\clearpage
\begin{figure}[p]\centering
\includegraphics[width=\textwidth,height=.96\textheight,keepaspectratio]{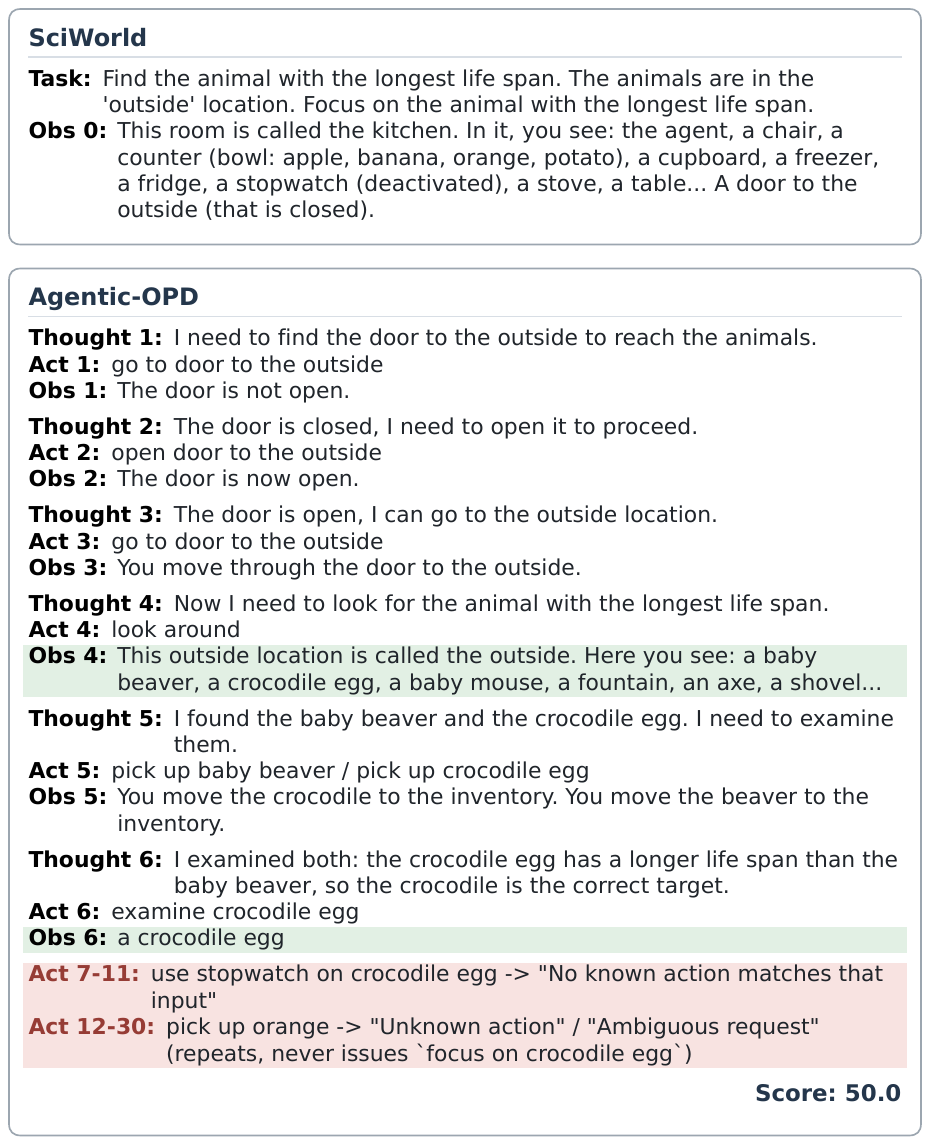}
\caption{\textbf{SciWorld case study.} The Agentic-OPD agent identifies the crocodile as the animal with the longest life span, but fails to issue the final focus action and receives partial reward.}
\label{fig:case-study-sciworld}
\end{figure}\clearpage
\begin{figure}[p]\centering
\includegraphics[width=\textwidth,height=.96\textheight,keepaspectratio]{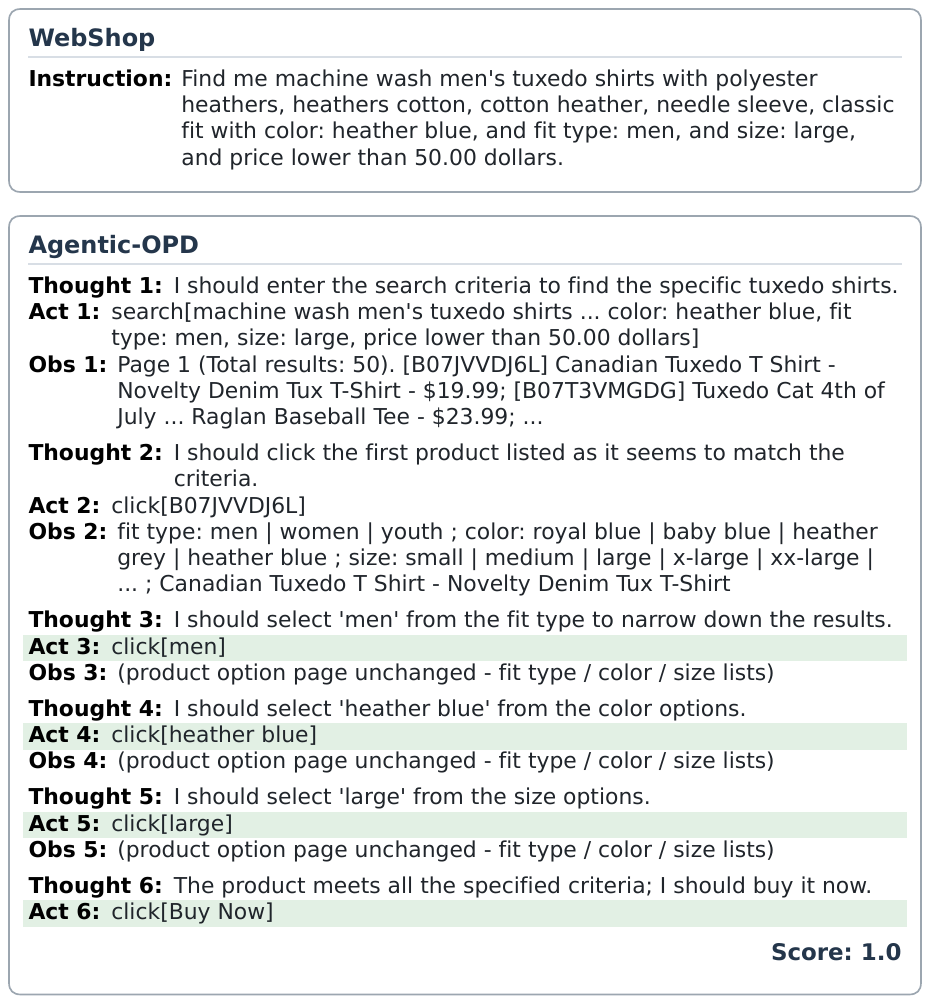}
\caption{\textbf{WebShop case study.} The agent searches for the requested product, selects the required attributes, and completes the purchase.}
\label{fig:case-study-webshop}
\end{figure}\clearpage
\begin{figure}[p]\centering
\includegraphics[width=\textwidth,height=.96\textheight,keepaspectratio]{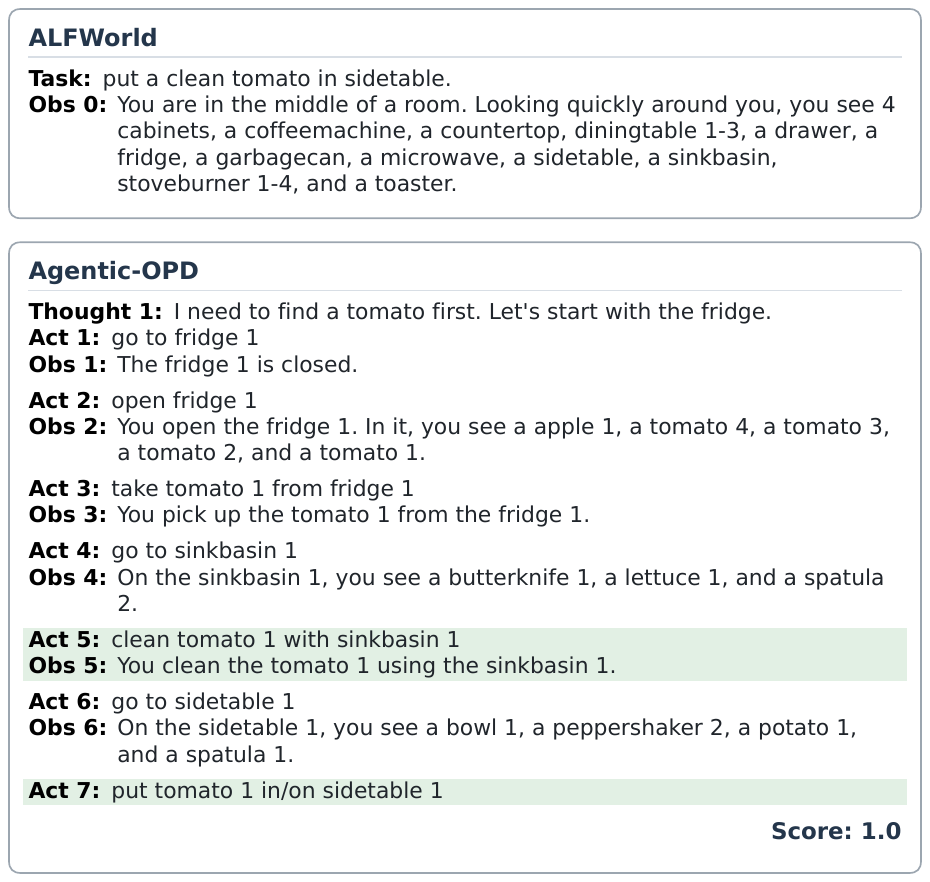}
\caption{\textbf{ALFWorld case study.} The agent locates and cleans a tomato before placing it on the sidetable, completing the task.}
\label{fig:case-study-alfworld}
\end{figure}\clearpage
\begin{figure}[p]\centering
\includegraphics[width=\textwidth,height=.96\textheight,keepaspectratio]{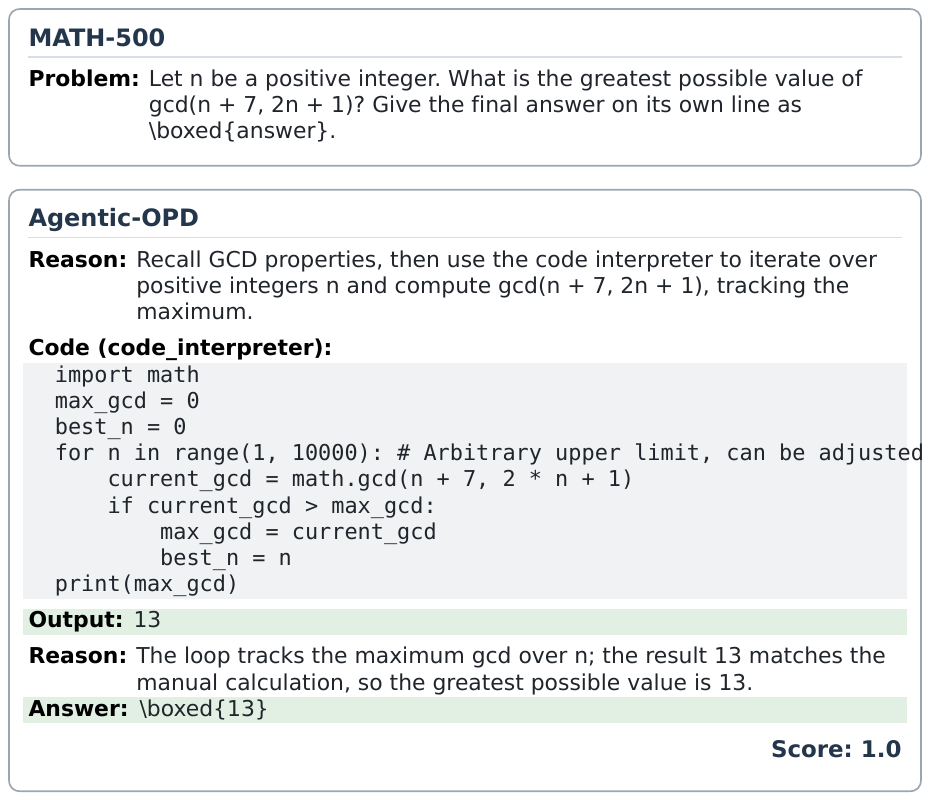}
\caption{\textbf{MATH-500 case study.} The agent uses the code interpreter to verify the maximum GCD and returns $\boxed{13}$.}
\label{fig:case-study-math500}
\end{figure}\clearpage
\begin{figure}[p]\centering
\includegraphics[width=\textwidth,height=.96\textheight,keepaspectratio]{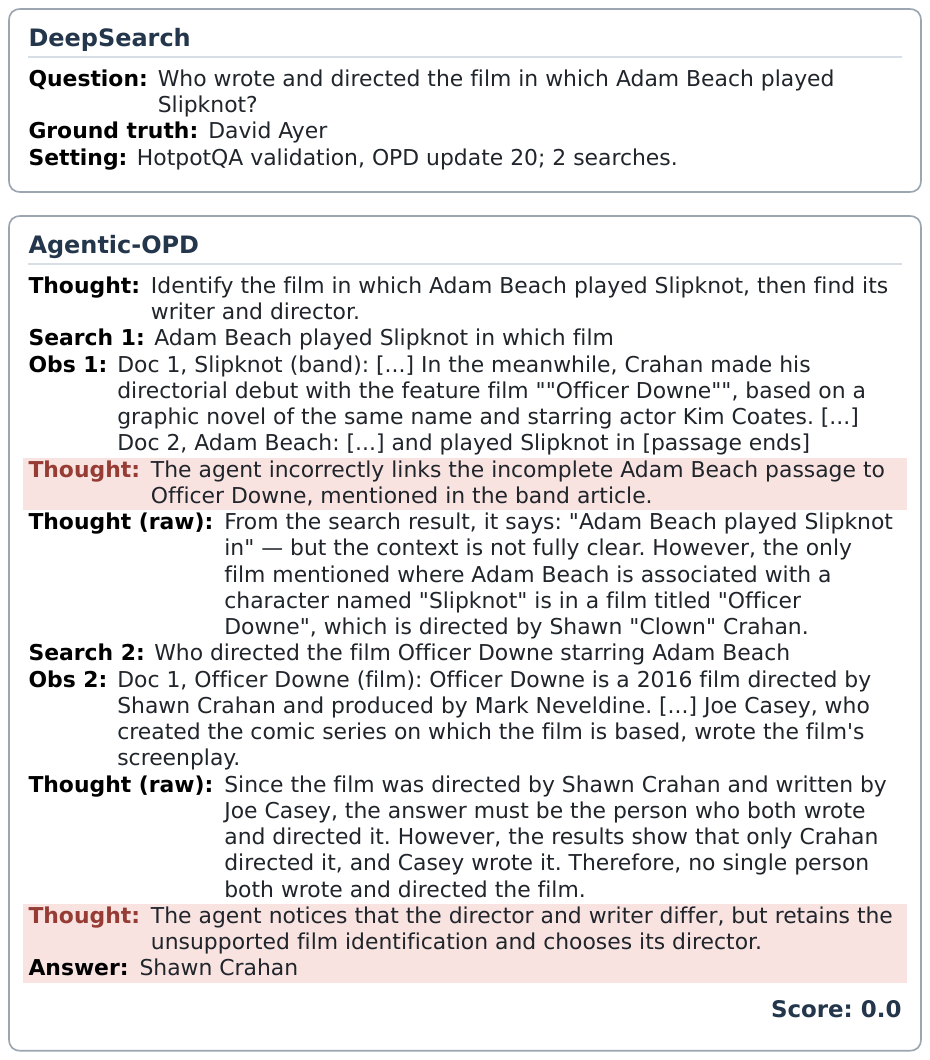}
\caption{\textbf{DeepSearch case study (incorrect).} At OPD update 20, the agent confuses the character Slipknot with the band, follows an unsupported link to \emph{Officer Downe}, and answers Shawn Crahan instead of David Ayer (score 0.0).}
\label{fig:case-study-deepsearch-incorrect}
\end{figure}\clearpage
\begin{figure}[p]\centering
\includegraphics[width=\textwidth,height=.96\textheight,keepaspectratio]{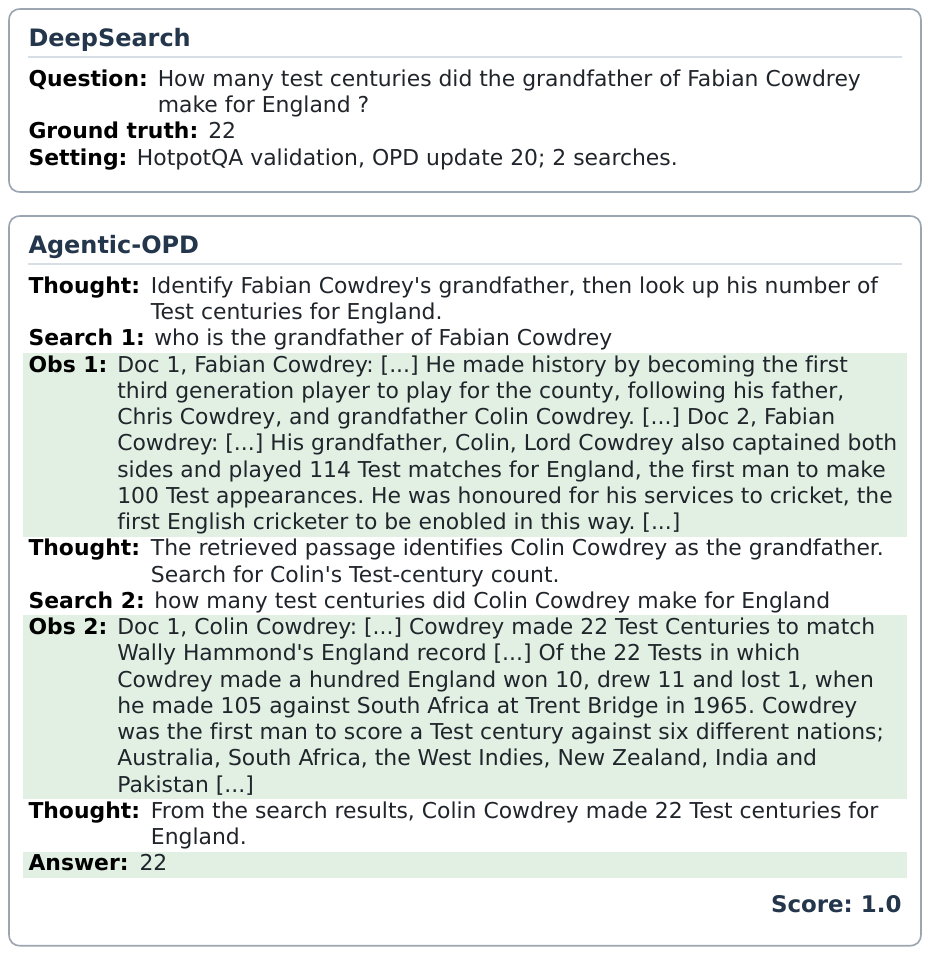}
\caption{\textbf{DeepSearch case study (correct).} At OPD update 20, the agent identifies Colin Cowdrey as Fabian Cowdrey's grandfather, then retrieves his 22 Test centuries for England and returns the correct answer (score 1.0).}
\label{fig:case-study-deepsearch-correct}
\end{figure}\clearpage

\end{document}